\documentclass{article}
\usepackage{spconf,amsmath,epsfig,amssymb}
\usepackage{multirow}
\usepackage{booktabs}
\usepackage{threeparttable}
\usepackage{color}
\usepackage{pifont}
\definecolor{gray}{rgb}{0.35,0.35,0.35}
\definecolor{blue}{rgb}{0,0,1}
\definecolor{red}{rgb}{1,0,0}
\definecolor{orange}{rgb}{0.75, 0.4, 0}
\definecolor{purple}{rgb}{0.5, 0.0, 0.5}

\begin{document}\sloppy

% Example definitions.
% --------------------
\def\x{{\mathbf x}}
\def\L{{\cal L}}

% Title.
% ------
\title{MixTConv: Mixed Temporal Convolutional Kernels for Efficient Action Recogntion}
%
% Single address.
% ---------------,
\name{Kaiyu Shan\textsuperscript{\rm 1}, Yongtao Wang\textsuperscript{\rm 1}, Zhuoying Wang\textsuperscript{\rm 1}, Tingting Liang\textsuperscript{\rm 1}, Zhi Tang\textsuperscript{\rm 1}, Ying Chen\textsuperscript{\rm 2}, Yangyan Li\textsuperscript{\rm 2}}
%Address and e-mail should NOT be added in the submission paper. They should be present only in the camera ready paper. 
\address{\textsuperscript{\rm 1}Wangxuan Institute of Computer Technology, Peking University\\
 \textsuperscript{\rm 2}Alibaba Cloud Intelligence Business Group\\
 \{shankyle, wyt, wzypku, tingtingliang, tangzhi\}@pku.edu.cn\\
  \{chenying.ailab, yangyan.lyy\}@alibaba-inc.com\\ 
}
%Address and e-mail should NOT be added in the submission paper. They should be present only in the camera ready paper. 

\maketitle

\begin{abstract}
To efficiently extract spatiotemporal features of video for action recognition, most state-of-the-art methods integrate 1D temporal convolution into a conventional 2D CNN backbone. However, they all exploit 1D temporal convolution of fixed kernel size (i.e., 3) in the network building block, thus have suboptimal temporal modeling capability to handle both long-term and short-term actions. To address this problem, we first investigate the impacts of different kernel sizes for the 1D temporal convolutional filters. Then, we propose a simple yet efficient operation called Mixed Temporal Convolution (MixTConv), which consists of multiple depthwise 1D convolutional filters with different kernel sizes. By plugging MixTConv into the conventional 2D CNN backbone ResNet-50, we further propose an efficient and effective network architecture named MSTNet for action recognition, and achieve state-of-the-art results on multiple benchmarks.
\end{abstract}


\section{Introduction}
\label{sec:intro}
Action recognition, which aims at assigning corresponding labels to the given videos, is a fundamental task for many real-world applications such as human-computer interaction and urban security systems. Temporal information is very important for action recognition. For example, it is hard to distinguish between “pulling something right to left” and “pulling something left to right” without temporal information\cite{GoyalKMMWKHFYMH17}. Hence, how to model both spatial and temporal information of video, i.e., extracting spatiotemporal features of video, is crucial for action recognition.

2D CNN-based action recognition approaches\cite{DBLP:conf/cvpr/KarpathyTSLSF14,wang2018temporal,DBLP:conf/eccv/ZhouAOT18,DBLP:conf/cvpr/FeichtenhoferPZ16} individually extract spatial features on sampled frames, which is efficient but struggle with temporal information modeling. On the contrast, 3D CNN-based methods\cite{DBLP:conf/cvpr/CarreiraZ17,DBLP:conf/iccv/TranBFTP15} jointly learn spatiotemporal features and achieve higher recognition accuracy, but bring huge computational cost. 
To address these problems, most state-of-the-art methods\cite{DBLP:conf/eccv/XieSHTM18,DBLP:conf/cvpr/TranWTRLP18,DBLP:conf/iccv/QiuYM17} integrate 1D temporal convolution into conventional 2D CNN to achieve good trade-off between efficiency and accuracy. Despite their success, their ability of modeling both long-term and short-term actions is not optimal since they all exploit 1D temporal convolution of fixed kernel size (i.e., 3) in the network building block. Moreover, they employ ordinary 1D convolution along temporal dimension, thus their efficiency can be further improved by depthwise convolution.

To tackle the issues mentioned above, we first study different lightweight temporal convolution methods and the impact of different kernel sizes along the temporal dimension. And we find that: 1) depthwise 1D convolution performs better than ordinary 1D convolution with large computation saving; 2) large kernel size of depthwise 1D convolution for temporal modeling does not always bring higher accuracy; 3) utilizing both large kernel and small kernel along the temporal dimension can capture long-term and short-term temporal information simultaneously, leading to better accuracy and efficiency. 

According to these findings, we propose a simple yet efficient temporal operation for spatiotemporal feature extraction, i.e., \textit{Mixed Temporal Convolution} (\textbf{MixTConv}). This operation partitions input channels into groups and performs depthwise 1D convolution with different kernel sizes to each group, such that it can extract temporal features of different scales. It has several superiorities: 1) compared with the 3D convolution and ordinary 1D convolution, it enjoys both higher accuracy and efficiency by extract multi-scale temporal features; 2) it keeps the same size between input and output, thus it is plug-in-play and can be flexibly inserted into any 2D CNNs. For instance, we insert the proposed MixTConv into the residual block of ResNet50\cite{DBLP:conf/cvpr/HeZRS16} to build a \textit{Mixed Spatiotemporal Network} (\textbf{MSTNet}) for action recognition, and experimental results on various large-scale public datasets show that: 1) the proposed MixTConv operation can significantly improve recognition accuracy of 2D CNN baseline by 27.6\% (from 20.5\% to 48.1\%) on Something-Something v1 and 31.4\% (from 30.4 to 61.8\%) on Something-Something v2; 2) compared with state-of-the-art method TSM\cite{lin2019tsm}, MSTNet obtains considerable recognition accuracy improvement more than 1.1\% with negligible parameter and computational cost.

To sum up, the contributions of this work are threefold:
\begin{itemize}
\item We propose a novel Mixed Temporal Convolution operation (MixTConv) for efficient spatiotemporal feature extraction, which mixes up multiple depthwise 1D convolutional filters with different kernel sizes.
\item We further propose a very efficient and effective network architecture named MSTNet for action recognition by plugging MixTConv into conventional 2D CNN, which significantly improves recognition accuracy of the 2D CNN baseline.
\item We achieve state-of-the-art results on multiple benchmarks, including Something-Something v1, Something-Something v2 and Jester.
\end{itemize}
\section{Related Work}
\label{sec:abstract}
\noindent\textbf{Spatiotemporal modeling}\quad
2D CNN-based methods\cite{wang2018temporal,DBLP:conf/cvpr/FeichtenhoferPZ16} directly use frame-wise prediction aggregation. For example, Simonyan \textit{et al.}\cite{DBLP:conf/cvpr/FeichtenhoferPZ16} designs a two-steam CNNs network by combining RGB input and optical flow results. TSN\cite{wang2018temporal} divides the video into N segments and samples one frame from each segment, then consensus the result by averaging. Despite their high efficiency, 2D CNN-based methods perform poorly on the action videos due to their weakness of temporal modeling. 3D CNN-based methods\cite{ji20133d,DBLP:conf/iccv/TranBFTP15,DBLP:conf/cvpr/CarreiraZ17} jointly learn spatiotemporal features in an elegant way. Tran \textit{et al.}\cite{DBLP:conf/iccv/TranBFTP15} proposes C3D based on VGG backbone\cite{simonyan2014very} to capture temporal features from a frame sequence. I3D\cite{DBLP:conf/cvpr/CarreiraZ17} introduces a 3D ConvNet based on 2D ConvNet inflation by expanding the filters and pooling kernels in an Inception V1 model\cite{szegedy2015going} into 3D convolutional kernels, so that it can leverage 2D network architecture designed for image classification and even their parameter weights pre-trained on ImageNet\cite{deng2009imagenet}. However, due to the model complexity, the pure 3D convolutional networks are resource-costly and prone to overfit\cite{DBLP:conf/eccv/XieSHTM18}. Hence, several methods focus on decomposing the 3D convolutions into separate 2D spatial and 1D temporal filters\cite{DBLP:conf/eccv/XieSHTM18,DBLP:conf/cvpr/TranWTRLP18,DBLP:conf/iccv/QiuYM17}. However, these methods still suffer from computational cost due to usage of ordinary 1D convolution.

\noindent\quad \textbf{Efficient operations and modules for temporal modeling}\quad 
Some methods attempt to trade off performance and computation by proposing different modules or temporal operations\cite{hussein2019timeception,DBLP:conf/eccv/ZhouAOT18,lin2019tsm}. TRN\cite{DBLP:conf/eccv/ZhouAOT18} adds temporal fusion after feature extraction, leading to limited improvement of performance. TSM utilizes \textit{shifting operation} which shifts 
a portion of the channels along the temporal dimension. Essentially, this operation is a fixed weight depthwise 1D convolution, which is not flexible enough for temporal modeling. 
Timeception\cite{hussein2019timeception} is another module which uses depthwise 1D convolution with different kernel sizes. However, it is quite different from our MixTConv with a more complex structure. Specifically, a Timeception layer divides input channels into several groups, and each group consists of multiple branches. Concretely, each branch is composed of a depthwise 1D convolution with different kernel sizes and a following 2D convolution layer. In contrast, our MixTConv is much more efficient: we divide the input channels into multiple groups and perform depthwise 1D convolution with different kernel sizes on each group. 
Moreover, the way of \cite{hussein2019timeception} to integrate the module is different from ours. In the network of \cite{hussein2019timeception}, four Timeception layers are stacked on top of the last convolution layer of a 3D CNN or 2D CNN, which is the late fusion as the same as TRN. As a contrast, MixTConv is inserted into all the blocks of the 2D CNN backbone to build our MSTNet.

\noindent\quad \textbf{Mixed Convolution} \quad
MixConv\cite{tan2019mixconv} uses 2D spatial convolution filters of different kernel sizes to extract spatial features of various resolutions, for improving image recognition accuracy. Differently, our MixTConv mixes up multiple depthwise 1D convolutional filters with different kernel sizes to capture both long-term and
short-term temporal information, for boosting action recognition performance.
\section{Methodology}
In this section, we first introduce the proposed Mixed Temporal Convolution operation(MixTConv) in sec ~\ref{sec3.1}. Then, the Mixed Spatiotemporal Block (MST Block) which intergrates MixTConv into 2D residual block is presented in sec ~\ref{sec3.2}. Finally, our proposed video recognition network Mixed Spatiotemporal Network (MSTNet) is introduced in sec ~\ref{sec3.3}.

\begin{figure*}[t]
\centering
\includegraphics[scale=0.5]{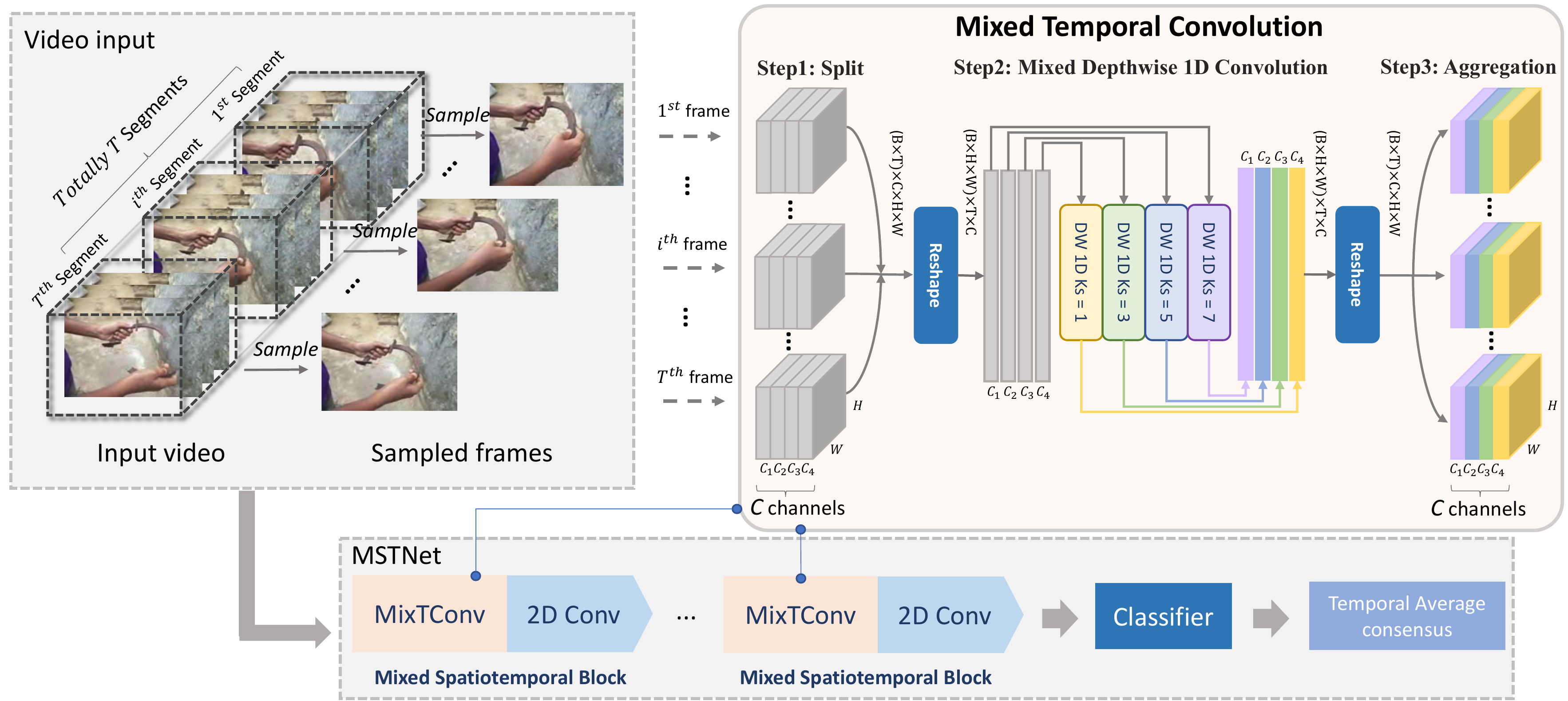}
\caption{The pipeline of the proposed video action recognition network Mixed Spatiotemporal Network(\textit{MSTNet}), base on the Mixed Temporal Convolution. "Ks" means kernel size, and "DW" means depthwise.}
\label{fig:pipeline}
\end{figure*}

\begin{figure}[t]
\centering
\includegraphics[scale=0.33]{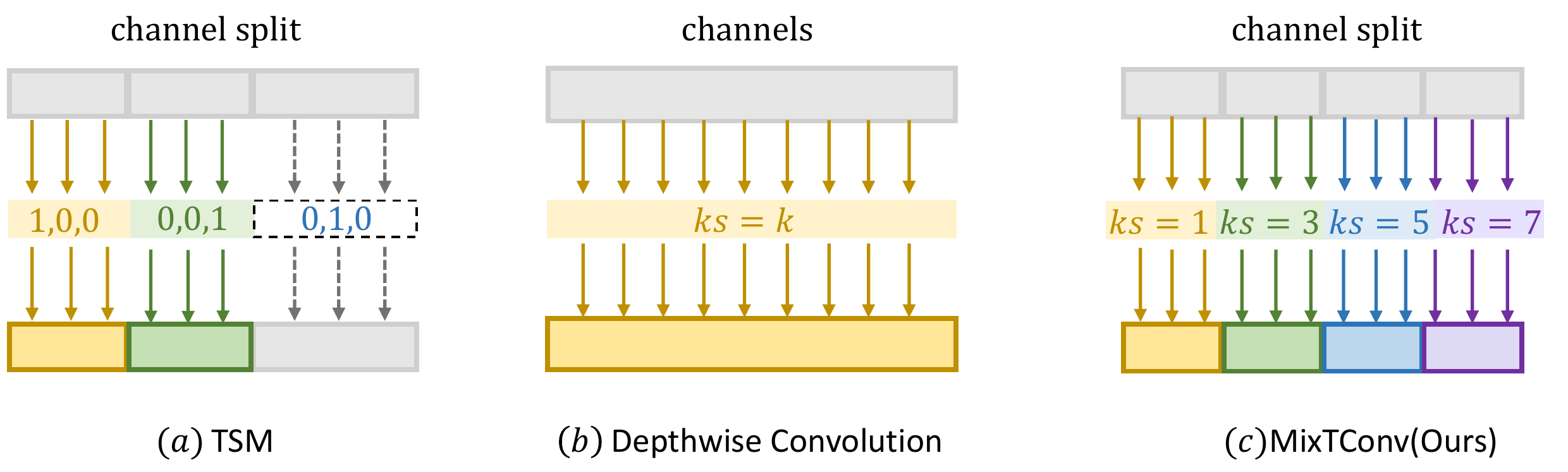}
\caption{Comparison of different temporal operations. (a)\textit{shift} temporal operation with \textit{fixed kernel weight and kernel size}. (b) \textit{learnable} temporal operation with the \textit{fixed kernel size} of depthwise 1D convolution. (c) \textit{Mixed Temporal Convolution}(MixTConv) with different kernel sizes of depthwise 1D convolution.}
\label{fig:compare}
\end{figure}

\begin{figure}[t]
\centering
\includegraphics[scale=0.25]{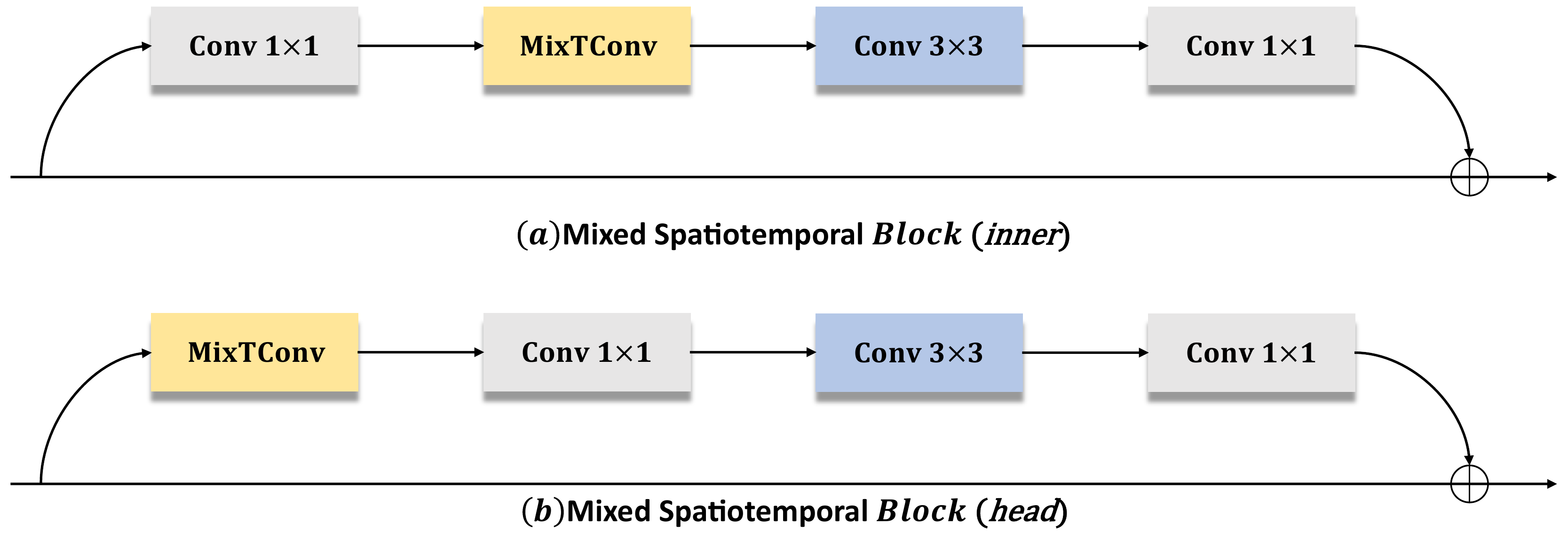}
\caption{Comparision for MST Block \textit{head} and MST Block \textit{inner}.}
\label{fig:block}
\end{figure}

\subsection{MixTConv: Mixed Temporal Convolution}
\label{sec3.1}
MixTConv is designed for \textit{efficient} and \textit{effective} temporal modeling. To achieve that, it has three engaging properties: 1) Unlike 3D CNNs which simultaneously convolve the spatial and temporal dimensions, our proposed MixTConv models these subspaces separately by decomposing the spatiotemporal modeling, and focus on temporal modeling; 2) Unlike the existing 2+1D methods that use ordinary 1D convolution, MixTConv applies depthwise 1D convolution, which significantly reduces the computation by a factor of C, where C is the number of input channels.
3) The depthwise fashion allows us to mix multiple depthwise 1D convolutions with different kernel sizes, thus can extract multi-scale temporal features and significantly boost the performance from the baseline with minor additional computation.

Here and after, We denote the input feature map for MixTConv operation as $F \in \mathbb{R}^{(B \times T) \times C \times H \times W}$, where $B$, $H$, $W$, $T$, $C$ is the batch size, height, weight, number of sampled frames and channel size, respectively.
As illustrated in Figure~\ref{fig:pipeline}, we firstly reshape 
$F$ as: $F \in \mathbb{R}^{(B \times H\times W) \times C \times T}$, 
and then apply the depthwise 1D convolution with $g$ different kernel sizes \{$k_{1}, ..., k_{g}$\} on the temporal dimension. Let $W_{m}$ denotes a depthwise 1D convolutional kernel with kernel size of $k_{m}$. Unlike vanilla depthwise convolution, MixTConv partitions channels into $g$ groups \{$\hat{F}^{1}, ..., \hat{F}^{g}$\} and applies depthwise 1D convolution with different kernel sizes to each group, where $c_{m}$ denotes channels in the $m$-$th$ group. Formally, the mixed 1D convolution is defined as: 

\begin{equation}
\hat{Z}_{i,t}^{m}=\sum_{j}\hat{F}_{t+j}^{i}W_{\frac{k_m -1}{2}+j},m=1,...,g
\end{equation}
Where $j\in[-\frac{k_{m}-1}{2},\frac{k_{m}-1}{2}]$ and $\hat{Z}_{i,t}^{m}$ is the value of $\hat{Z}^{m}$ at the $t$-$th$ frame and $i$-$th$ channel.

The final output tensor is a concatenation of all the output tensor \{$\hat{Z}^{1},...,\hat{Z}^{g}$\} :
\begin{equation}
Z = Concat(\hat{Z}^{1},...,\hat{Z}^{g}).
\end{equation}

\textbf{Discussion}
Our method is related to the current state-of-the-art method TSM\cite{lin2019tsm}. In fact, the \textit{shift} operation is a special case of our proposed MixTConv, more specifically, equal to a \textit{fixed weight} depthwise 1D convolution with \textit{fixed kernel size} of 3, where temporal kernel is fixed as: [0, 1, 0] for \textit{static channels}(3/4 of total channels), [1, 0, 0] for \textit{backward-shift channels}(1/8 of total channels), and [0, 0, 1] for \textit{forward-shift channels}(1/8 of total channels), shown in Figure~\ref{fig:compare}(a). 
Our experiment shows that, using depthwise 1D convolution with \textit{learnable weight} (Figure~\ref{fig:compare}(b)) and \textit{multiple kernel sizes} (Figure~\ref{fig:compare}(c)) along temporal dimension is more effective than these hand-crafted temporal kernels to capture pyramidal temporal contextual information.

\subsection{Mixed Spatiotemporal Block}
\label{sec3.2}
Our proposed MixTConv can be flexibly plugged into any existing 2D architectures with limited computational cost, thus can extract spatiotemporal features efficiently. 
As illustrated in Figure~\ref{fig:block}, taking ResNet block\cite{DBLP:conf/cvpr/HeZRS16} as an example, a straight-forward way to apply MixTConv is to plug it after the first $1 \times 1$ convolution, which is denoted as MST Block \textit{inner}(Figure~\ref{fig:block}(a)). However, it harms the capability of spatiotemporal feature learning since the channels are reduced in the bottleneck. To address such issue, we propose MST Block \textit{head}(Figure~\ref{fig:block}(b)), which plugs MixTConv between residual operation and the first $1 \times 1$ convolution. The computational cost is negligible for both MST Block \textit{head} and MST Block \textit{inner}(0.18G FLOPS and 0.05G FLOPS) due to the depthwise fashion. 
As shown in Table ~\ref{tab:place}, MST Block \textit{head} achieves better recognition accuracy, verifying our assumption. So that we use MST Block \textit{head} as our MST Block.
\begin{table}[t]
\caption{Comparisons between the proposed MSTNet and 2D CNN baseline TSN(protocol: ResNet-50 8f input, 2 clips for all datasets, full-resolution).} \label{tab:compare}
% \vspace{-6pt}
\footnotesize
\begin{center}
\label{tab:compare_2d}
\begin{tabular}{c|c|c|c|c|c}
\toprule
\textbf{Dataset} &  \textbf{Model} &\textbf{MixTConv} &  \textbf{Top-1} &  \textbf{Top-5} &  \textbf{$\Delta$ Top-1}  \\ 
\hline

 \multirow{2}{*}{\shortstack{Something\\v1}}  & TSN\cite{wang2018temporal} &\ding{56}& 20.5  & 47.5 & \multirow{2}{*}{+27.6} \\ 
  & Ours &\ding{51}& \textbf{48.1} & \textbf{77.3} & \\ \cmidrule(l){1-6}
   \multirow{2}{*}{\shortstack{Something\\v2}}  & TSN\cite{wang2018temporal} &\ding{56} & 30.4  & 61.0 & \multirow{2}{*}{+31.4} \\ 
  & Ours &\ding{51}& \textbf{61.8} & \textbf{87.8} & \\ \cmidrule(l){1-6}
   \multirow{2}{*}{\shortstack{Jester}}  & TSN\cite{wang2018temporal} &\ding{56}& 83.9  & 99.6 & \multirow{2}{*}{+13.0} \\ 
  & Ours &\ding{51}& \textbf{96.9} & \textbf{99.9} & \\
\bottomrule
\end{tabular}
\end{center}
% \vspace{-15pt}
\end{table}
\subsection{Network Design}
\label{sec3.3}
Based on the \textit{Mixed Spatiotemporal Block}, a network named \textit{MSTNet} is built for action recognition. 
In order to keep the framework efficient, we choose the 2D ResNet50\cite{DBLP:conf/cvpr/HeZRS16} as our backbone to achieve a good trade-off between the accuracy and the speed. We replace all residual blocks with the proposed MST Blocks. The pipeline follows the popular TSN\cite{wang2018temporal} framework, which samples frames sparsely and then passes them through the 2D CNNs followed by a consensus aggregation function(e.g. Average pooling). For both TSN and MSTNet, the final output of a video is:

\begin{table}[t]
\scriptsize
\centering
\caption{Comparisons of different temporal operations and configurations (i.e., the kernel size and the combinations of the filters) on Something-Something v1. 
% The results are based on 8 frames, one clip and one center crop.
"ks"denotes kernel size and * denotes shifting convolution.}
\begin{tabular}{lccccc}
\toprule
\textbf{Method}& \textbf{Kernel Size}& \textbf{Dilation}&\textbf{Learnable}&  \textbf{Top-1} & \textbf{FLOPS}\\
\hline
TSN(baseline)\cite{wang2018temporal}& -& -& \ding{56}& 19.7 & 33G\\
TSN+Ordinary 1D & 3& 1& \ding{51} & 41.0&43G \\
TSM*\cite{lin2019tsm}& 3*& 1& \ding{56}&45.6 & 33G\\
TSN+ks3& 3& 1& \ding{51}&45.9 & 33.13G\\
TSN+ks5& 5& 1& \ding{51}&46.3 & 33.23G\\
TSN+ks7& 7& 1& \ding{51}&45.8 & 33.32G\\
TSN+ks13& 1,3& 1& \ding{51}&45.8 & 33.09G\\
TSN+ks135& 1,3,5& 1& \ding{51}&46.4 & 33.13G\\
TSN+ks1357& 1,3,5,7& 1& \ding{51}&\textbf{46.7} & 33.18G\\
TSN+ks357& 3& 1,2,3& \ding{51}&46.4 & 33.13G\\
\bottomrule
\end{tabular}
\label{tab:ks}
\end{table}

\begin{table}[t]
\small
\centering
\caption{Comparisons of two blocks that integrate MixTConv on Something-Something v1. 
% The results are based on 8 frames, one clip and one center crop. 
}
\begin{tabular}{lccccc}
\toprule
\textbf{Method}& \textbf{Insert place}&  \textbf{Top-1} & \textbf{Top-5} & \textbf{FLOPS}\\
\hline
MST Block \textit{inner}& after 1x1& 45.8&74.4 & 33.05G\\
MST Block \textit{head}& before 1x1& \textbf{46.7}& 75.6 & 33.18G\\
\bottomrule
\end{tabular}
\label{tab:place}
\end{table}

\begin{equation}
S = avg(\hat{S}^{i}),   \quad i \in 0,1,...,T,
\end{equation}
where T is the number of sampled frames(segments) in the video, and $\hat{S}^{i}$ is the output feature of the $i$-$th$ frame by the network.
It is obvious that, by using simple consensus aggregation on final score of each frame, TSN lacks capability of modeling the temporal relationship. Results in Table ~\ref{tab:compare_2d} shows that, with MixTConv operation, MSTNet significantly boosts the performance of TSN.

\section{Experiment}
\subsection{Dataset and Implementation details}
\label{implement}
\textbf{Dataset} \textit{Something-Something v1} and \textit{v2} \cite{GoyalKMMWKHFYMH17} are two large-scale video datasets for action recognition. 
The datasets contain 110k(v1) and 220k(v2) videos, respectively, each with around 50 frames. The videos are annotated into 174 fine-grained human action classes with various objects and viewpoints. In these two datasets, videos with similar labels, e.g. "\textit{opening the door}" v.s. "\textit{closing the door}", are indistinguishable without exploiting temporal information. \textit{Jester}\cite{materzynska2019jester} is a large collection of densely-labeled video clips that show humans performing pre-definded hand gestures in front of a laptop camera or webcam.
It contains 27 classes of hand gestures and each with around 5,000 instances, which makes it possible to train robust model on gesture recognition.

\textbf{Training} 
All experiments in this paper adopt ResNet-50\cite{DBLP:conf/cvpr/HeZRS16} pre-trained on ImageNet\cite{deng2009imagenet} as backbone, and further fine-tuned on target datasets. 
For an apple-to-apple comparison with state-of-the-art methods\cite{lin2019tsm}, we strictly follow the same training protocols. The initial learning rate is set as 0.01 and decays by 0.1 at epoch 30\&40\&45. We train the networks for 50 epochs with weight decay of 5e-4, batch size of 64 and dropout as 0.5.
For data augmentation, we follow TSN\cite{wang2018temporal} to sample one frame from every 8 or 16 segments. Then we resize their short side to 256 and meanwhile keep the aspect ratio as 4:3. After that, we exploit corner cropping and scale-jittering. 

\begin{table*}[t]
\centering
\scriptsize
\caption{Comparisons with state-of-the-art methods on Something-Something v1 and Something-Something v2.}
\begin{tabular}{c|c|c|c|c|c|c|c|c|cc}
\toprule

\multirow{2}{*}{\textbf{Method}}& \multirow{2}{*}{\textbf{Backbone}}& \multirow{2}{*}{\textbf{Modality}}& \multirow{2}{*}{\textbf{Frames}}& \multirow{2}{*}{\textbf{Params}}& \multirow{2}{*}{\textbf{FLOPs}}& \multicolumn{2}{c|}{Something-Something v1}& \multicolumn{2}{c}{Something-Something v2}\tabularnewline
\cline{7-10}
& & & & & & \textbf{Val Top-1}& \textbf{Val Top-5}& \textbf{Val Top-1}& \textbf{Val Top-5}\\
\hline
TSN\cite{wang2018temporal}ECCV'16& BNIception& RGB& 8& 10.7M& 16G& 19.5 & - &-&-\\
TSN(baseline)\cite{wang2018temporal}ECCV'16& ResNet-50& RGB&8&24.3M& 33G& 19.7& 46.6 &27.8&57.6\\
TRN Multiscale\cite{DBLP:conf/eccv/ZhouAOT18}ECCV'18& BNInception& RGB& 8& 18.3M& 16G& 34.4& -&44.8&77.6\\
TRN Two-steam\cite{DBLP:conf/eccv/ZhouAOT18}ECCV'18& BNInception& RGB+Flow& 8+8& 36.6M& -& 42.0& -&55.5&83.1\\
\hline
I3D\cite{DBLP:conf/cvpr/CarreiraZ17}CVPR'17& 3D ResNet-50& RGB& 32$\times$2clips& 28.0M& 153G$\times$2& 41.6& 72.2&-&-\\
NL*+I3D\cite{wang2018non}CVPR'18& 3D ResNet-50& RGB& 32$\times$2clips& 35.3M& 168G$\times$2& 44.4& 76.0&-&-\\
NL*+I3D+GCN\cite{wang2018videos}ECCV'18& 3D ResNet-50+GCN& RGB& 32$\times$2clips& 62.2M& 303G$\times$2& 46.1& 76.8&-&-\\
\hline
ECO\cite{DBLP:conf/eccv/ZolfaghariSB18}ECCV'18& BNIn$\rm{c^{*}}$+Res3D$\rm{18^{*}}$ & RGB& 8& 47.5M& 32G& 39.6& -&-&-\\
ECO\cite{DBLP:conf/eccv/ZolfaghariSB18}ECCV'18& BNIn$\rm{c^{*}}$+Res3D$\rm{18^{*}}$ & RGB& 16& 47.5M& 64G& 41.4& -&-&-\\
$\rm ECO_{En}\emph{Lite}$\cite{DBLP:conf/eccv/ZolfaghariSB18}ECCV'18& BNIn$\rm{c^{*}}$+Res3D$\rm{18^{*}}$ & RGB& 92& 150M& 267G& 46.4& -&-&-\\
\hline
TSM\cite{lin2019tsm}ICCV'19& ResNet-50& RGB& 8& 24.3M& 33G& 45.6& 74.2&58.7*&85.4\\
TSM\cite{lin2019tsm}ICCV'19& ResNet-50& RGB& 16& 24.3M& 65G& 47.2& 77.1&61.0*&86.8\\
\hline
\hline
\small\textit{Ours:} & & & & & & & &\quad \\
%\textbf{PstNet-inner(Ours)}& ResNet-50& 8& 25.8M& & 44.2& \\
\textbf{MSTNet}& ResNet-50& RGB& 8& 24.3M& 33.2G& 46.7& 75.4&59.5&86.0\\
\textbf{MSTNet}& ResNet-50& RGB& 16& 24.3M& 65.3G& \textbf{48.4}& \textbf{78.8}&\textbf{61.8}&\textbf{87.3}\\

\bottomrule
\end{tabular}
\label{tab:ssv1}
\begin{tablenotes}
  \footnotesize
    \item $^{\rm *}$BNInc means BNInception, $^{\rm *}$Res3D18 means 3D Resnet 18, $^{\rm *}$NL means Non-Local\cite{wang2018non}.
    \item $^{\rm *}$Using offical released pre-trained weight and testing with one clip and center crop.
\end{tablenotes}
\end{table*}

\begin{table}[t]
\footnotesize
\centering
\caption{Comparison of state-of-the-art methods on Jester.}
\begin{tabular}{lccccc}
\toprule
\textbf{Method}& \textbf{Modality}& \textbf{Frames}&\textbf{FLOPS}& \textbf{Top-1}& \textbf{Top-5}\\
\hline
TSN\cite{wang2018temporal}& RGB& 8& 33G& 81.0 & 99.0 \\
TSN\cite{wang2018temporal}& RGB&16& 65G& 82.3& 99.2\\
TRN-MS*\cite{DBLP:conf/eccv/ZhouAOT18}& RGB&8& 16G&93.7& -\\
\hline
TSM*\cite{lin2019tsm}& RGB& 8&  33G&94.5*&99.7 \\
TSM*\cite{lin2019tsm}& RGB& 16&   65G&95.3*& 99.8\\
\hline
\small\textit{Ours:} & & & & & \quad \\
\textbf{MSTNet}& RGB& 8& 33.2G&  96.0& 99.8\\
\textbf{MSTNet}& RGB& 16& 65.3G&   \textbf{96.8}& \textbf{99.8} \\
\bottomrule
\end{tabular}
\label{tab:jester}
\begin{tablenotes}
  \footnotesize
    \item $^{\rm *}$Using offical released pre-trained weight and testing with one clip and center crop.
    \item $^{\rm *}$ MS means multi-scale.
\end{tablenotes}
\end{table}

\textbf{Testing}
 A common practice for testing is to apply 10 crops to each
frame\cite{wang2018temporal,DBLP:conf/cvpr/CarreiraZ17}. Moreover, many state-of-the-art methods\cite{ji20133d,DBLP:conf/cvpr/TranWTRLP18} use dense frames of 64 or 128 in multiple clips(e.g. 10), leading to a huge
computation. For efficiency, we use only \textbf{one clip} per video and the center 224x224 crop for evaluation based on \textbf{RGB only} if not specified. Also, for direct comparison to 2D CNN baseline\cite{wang2018temporal}, we sample 2 clips per video and use the full resolution image with shorter side of 256 for evaluation(as in Table~\ref{tab:compare_2d}).

\subsection{Ablation Study}
\label{ablation study}
\textbf{Improving 2D CNN Baselines} MixTConv can be plugged into any normal 2D CNNs and boost their performance on video recognition. We verify its effectiveness by comparing the performance of MSTNet and the 2D CNN basline, TSN\cite{wang2018temporal}, with the same training and testing protocol. Noting that, the only difference between MSTNet and TSN is with or without MixTConv. Table~\ref{tab:compare} shows that 2D CNN baseline cannot achieve a good accuracy on the datasets with temporal information(i.e., Something-Something v1 and v2), but once equipped with MixTConv, the performance improves significantly with negligible increments of computational cost and parameters. These results demonstrate that MixConv is very effective and efficient for action recognition.

\textbf{Comparison of different temporal operations and configurations} We further compare temporal aggregation with different configurations (i.e., the kernel sizes and the combinations of the filters). 
As shown in Table~\ref{tab:ks}: 1)Depthwise 1D convolution achieves better performance than ordinary 1D with large computation saving; 2)Depthwise 1D convolution with kernel size = 3 performs better than \textit{shifting} operation, which implies that \textit{fixed weight} temporal convolution is not good enough for temporal modeling; 3)Larger kernel size along temporal axes not always lead to higher accuracy(TSN+ks7 gets 0.5 \% lower accuracy than that of TSN+ks5); 4)combination of multiple kernel sizes achieve much better performance than single kernel size, demonstrating that the design of MixTConv is effective and reasonable.

\textbf{Plugging Position} We further explore where to plug the MixTConv in ResNet building block. Table~\ref{tab:place} shows that MST Block \textit{head} performs better with nearly same computational complexity to MST Block \textit{inner}. We guess the reason is that depthwise fashion needs more channels to model features, which is proved in MobilenetV2\cite{sandler2018mobilenetv2}. Hence, we finally choose MST Block \textit{head} as the network building block.

\subsection{Comparison with the State-of-the-Art}
\label{compare}
\textbf{Something-Something v1} \quad We compare MSTNet with state-of-the-art methods on Something-Something v1 in Table~\ref{tab:ssv1}. The comparison details are as follows:
1) TRN\cite{DBLP:conf/eccv/ZhouAOT18} and TSN\cite{wang2018temporal} are based on 2D CNNs. TSN achieves poor performance due to the lack of temporal modeling. Notably, our single-stream network outperforms two-steam TRN\cite{DBLP:conf/eccv/ZhouAOT18} by 5\% absolutely, which implies the importance of temporal fusion for all layers. 
2) Non-local I3D\cite{wang2018non} with GCN\cite{wang2018videos} is the state-of-the-art 3D CNN based model. It's worth noting that, the GCN needs a Reion Proposal Network(RPN)\cite{ren2015faster} trained on other object detection datasets to get the bounding boxes, which has extra training cost. Compared with the Non-local I3D+GCN, our MSTNet achieves 0.8 \% better accuracy with 20$\times$ fewer FLOPs on the validation dataset.
3) ECO\cite{DBLP:conf/eccv/ZolfaghariSB18} and TSM\cite{lin2019tsm} are two state-of-the-art efficient action recognition methods. Compared to ECO, our method achieves 0.3\% better accuracy at 9$\times$ less computation with 6$\times$ less parameters. Compared to TSM, we achieve 1.1\% and 1.2\% better accuracy with little extra computational cost(0.005 \%). These results demonstrate that our proposed \textit{Mixed Temporal Convolution}(MixTConv) is a better way to model temporal information than other temporal operations like 3D convolution and shifting.

\textbf{Something-Something v2} \quad As illustrated in Table~\ref{tab:ssv1}, for this larger and newer dataset to the previous v1, our model also achieves better results than SOTA methods, with RGB modality only. These results demonstrate the effectiveness of the proposed MixTConv operation and MSTNet for action recognition once again. 

\textbf{Jester} \quad As shown in Table~\ref{tab:jester}, on the benchmark Jester, our MSTNet also gains a large improvement compared to the TSN baseline(+15\%), and outperforms all the recent state-of-the-art methods, for the task of gesture recognition.

\section{Conclusion}
In this work, we propose a lightweight and plug-and-play operation named Mixed Temporal Convolution (MixTConv) for action recognition, which partitions input channels into groups and performs depthwise 1D convolution with different kernel sizes to capture multi-scale temporal information. It can be flexibly inserted into any 2D CNN backbones to enable temporal modeling with negligible extra computational cost. We further design a Mixed Spatiotemporal Network (MSTNet) for action recognition, by plugging MixTConv into the building block of ResNet-50. Experimental results on Something-Something v1, v2 and Jester benchmarks consistently indicate the superiority of the proposed MSTNet with the MixTConv operation. Additional ablation studies further demonstrate that the designs of the proposed MixTConv operation and MSTNet are effective and reasonable.

\bibliographystyle{IEEEbib}
\bibliography{action}

\begin{thebibliography}{10}

\bibitem{GoyalKMMWKHFYMH17}
Raghav Goyal, Samira~Ebrahimi Kahou, and et~al.,
\newblock ``The "something something" video database for learning and
  evaluating visual common sense,''
\newblock in {\em {ICCV}}, 2017.

\bibitem{DBLP:conf/cvpr/KarpathyTSLSF14}
Andrej Karpathy, George Toderici, and et~al.,
\newblock ``Large-scale video classification with convolutional neural
  networks,''
\newblock in {\em {CVPR}}, 2014.

\bibitem{wang2018temporal}
Limin Wang, Yuanjun Xiong, and et~al.,
\newblock ``Temporal segment networks for action recognition in videos,''
\newblock {\em TPAMI}, 2018.

\bibitem{DBLP:conf/eccv/ZhouAOT18}
Bolei Zhou, Alex Andonian, and et~al.,
\newblock ``Temporal relational reasoning in videos,''
\newblock in {\em {ECCV}}, 2018.

\bibitem{DBLP:conf/cvpr/FeichtenhoferPZ16}
Christoph Feichtenhofer, Axel Pinz, and Andrew Zisserman,
\newblock ``Convolutional two-stream network fusion for video action
  recognition,''
\newblock in {\em {CVPR}}, 2016.

\bibitem{DBLP:conf/cvpr/CarreiraZ17}
Jo{\~{a}}o Carreira and Andrew Zisserman,
\newblock ``Quo vadis, action recognition? {A} new model and the kinetics
  dataset,''
\newblock in {\em {CVPR}}, 2017.

\bibitem{DBLP:conf/iccv/TranBFTP15}
Du~Tran, Lubomir~D. Bourdev, and et~al.,
\newblock ``Learning spatiotemporal features with 3d convolutional networks,''
\newblock in {\em {ICCV}}, 2015.

\bibitem{DBLP:conf/eccv/XieSHTM18}
Saining Xie, Chen Sun, and et~al.,
\newblock ``Rethinking spatiotemporal feature learning: Speed-accuracy
  trade-offs in video classification,''
\newblock in {\em {ECCV}}, 2018.

\bibitem{DBLP:conf/cvpr/TranWTRLP18}
Du~Tran, Heng Wang, and et~al.,
\newblock ``A closer look at spatiotemporal convolutions for action
  recognition,''
\newblock in {\em {CVPR}}, 2018.

\bibitem{DBLP:conf/iccv/QiuYM17}
Zhaofan Qiu, Ting Yao, and Tao Mei,
\newblock ``Learning spatio-temporal representation with pseudo-3d residual
  networks,''
\newblock in {\em {ICCV}}, 2017.

\bibitem{DBLP:conf/cvpr/HeZRS16}
Kaiming He, Xiangyu Zhang, and et~al.,
\newblock ``Deep residual learning for image recognition,''
\newblock in {\em CVPR}, 2016.

\bibitem{lin2019tsm}
Ji~Lin, Chuang Gan, and Song Han,
\newblock ``Tsm: Temporal shift module for efficient video understanding,''
\newblock in {\em ICCV}, 2019.

\bibitem{ji20133d}
S~Ji, Wei Xu, and et~al.,
\newblock ``3d convolutional neural networks for human action recognition,''
\newblock {\em TPAMI}, 2013.

\bibitem{simonyan2014very}
Karen Simonyan and Andrew Zisserman,
\newblock ``Very deep convolutional networks for large-scale image
  recognition,''
\newblock {\em ICLR}, 2014.

\bibitem{szegedy2015going}
Christian Szegedy, Wei Liu, and et~al.,
\newblock ``Going deeper with convolutions,''
\newblock in {\em CVPR}, 2015.

\bibitem{deng2009imagenet}
Jia Deng, Wei Dong, and et~al.,
\newblock ``Imagenet: A large-scale hierarchical image database,''
\newblock in {\em CVPR}, 2009.

\bibitem{hussein2019timeception}
Noureldien Hussein, Efstratios Gavves, and Arnold~WM Smeulders,
\newblock ``Timeception for complex action recognition,''
\newblock in {\em CVPR}, 2019.

\bibitem{tan2019mixconv}
Mingxing Tan and Quoc~V Le,
\newblock ``Mixconv: Mixed depthwise convolutional kernels,''
\newblock {\em ArXiv, abs/1907.09595}, vol. 7, 2019.

\bibitem{materzynska2019jester}
Joanna Materzynska, Guillaume Berger, Ingo Bax, and Roland Memisevic,
\newblock ``The jester dataset: A large-scale video dataset of human
  gestures,''
\newblock in {\em ICCVW}, 2019.

\bibitem{wang2018non}
Xiaolong Wang, Ross Girshick, and et~al.,
\newblock ``Non-local neural networks,''
\newblock in {\em CVPR}, 2018.

\bibitem{wang2018videos}
Xiaolong Wang and Abhinav Gupta,
\newblock ``Videos as space-time region graphs,''
\newblock in {\em ECCV}, 2018.

\bibitem{DBLP:conf/eccv/ZolfaghariSB18}
Mohammadreza Zolfaghari, Kamaljeet Singh, and Thomas Brox,
\newblock ``{ECO:} efficient convolutional network for online video
  understanding,''
\newblock in {\em {ECCV}}, 2018.

\bibitem{sandler2018mobilenetv2}
Mark Sandler, Andrew Howard, Menglong Zhu, Andrey Zhmoginov, and Liang-Chieh
  Chen,
\newblock ``Mobilenetv2: Inverted residuals and linear bottlenecks,''
\newblock in {\em CVPR}, 2018.

\bibitem{ren2015faster}
Shaoqing Ren, Kaiming He, Ross Girshick, and Jian Sun,
\newblock ``Faster r-cnn: Towards real-time object detection with region
  proposal networks,''
\newblock in {\em NIPS}, 2015.

\end{thebibliography}

\end{document}